\documentclass[letterpaper]{article} 
\usepackage{aaai2026}  
\usepackage{times}  
\usepackage{helvet}  
\usepackage{courier}  
\usepackage[hyphens]{url}  
\usepackage{graphicx} 
\usepackage{natbib}  
\usepackage{caption} 
\usepackage{algorithm}
\usepackage{algorithmic}

\usepackage{newfloat}
\usepackage{listings}
\DeclareCaptionStyle{ruled}{labelfont=normalfont,labelsep=colon,strut=off} 
\floatstyle{ruled}
\newfloat{listing}{tb}{lst}{}
\floatname{listing}{Listing}
\usepackage{multirow}
\usepackage{makecell}
\usepackage{booktabs}
\usepackage{amsfonts}
\newcommand{\figref}[1]{\mbox{Fig.~\ref{#1}}}
\newcommand{\tabref}[1]{\mbox{Tab.~\ref{#1}}}
\def\tilbf#1{\tilde{\mathbf{#1}}}
\def\mathbi#1{\textbf{\em #1}}
\newcolumntype{Y}{>{\centering\arraybackslash}X}
\newcolumntype{L}[1]{>{\raggedright\arraybackslash}p{#1}}
\newcolumntype{C}[1]{>{\centering\arraybackslash}p{#1}}
\newcolumntype{R}[1]{>{\raggedleft\arraybackslash}p{#1}}

\title{Object-Centric Framework for Video Moment Retrieval}
\author {
    Zongyao Li\textsuperscript{\rm 1},
    Yongkang Wong\textsuperscript{\rm 2},
    Satoshi Yamazaki\textsuperscript{\rm 1},
    Jianquan Liu\textsuperscript{\rm 1},
    Mohan Kankanhalli\textsuperscript{\rm 2}
}
\affiliations {
    \textsuperscript{\rm 1}Visual Intelligence Research Laboratories, NEC Corporation\\
    \textsuperscript{\rm 2}National University of Singapore\\
    zongyao-li@nec.com, yongkang.wong@nus.edu.sg, s-yamazaki31@nec.com, jqliu@nec.com, mohan@comp.nus.edu.sg
}

\usepackage{bibentry}

\begin{document}

\maketitle

\begin{abstract}
Most existing video moment retrieval methods rely on temporal sequences of frame- or clip-level features that primarily encode global visual and semantic information. However, such representations often fail to capture fine-grained object semantics and appearance, which are crucial for localizing moments described by object-oriented queries involving specific entities and their interactions. In particular, temporal dynamics at the object level have been largely overlooked, limiting the effectiveness of existing approaches in scenarios requiring detailed object-level reasoning. To address this limitation, we propose a novel object-centric framework for moment retrieval. Our method first extracts query-relevant objects using a scene graph parser and then generates scene graphs from video frames to represent these objects and their relationships. Based on the scene graphs, we construct object-level feature sequences that encode rich visual and semantic information. These sequences are processed by a relational tracklet transformer, which models spatio-temporal correlations among objects over time. By explicitly capturing object-level state changes, our framework enables more accurate localization of moments aligned with object-oriented queries. We evaluated our method on three benchmarks: Charades-STA, QVHighlights, and TACoS. Experimental results demonstrate that our method outperforms existing state-of-the-art methods across all benchmarks.
\end{abstract}


\section{Introduction}
\label{sec:1}

Video moment retrieval, also referred to as video temporal grounding, aims to localize temporal segments within a video that correspond to a given natural language query. Recent advances have primarily focused on enhancing cross-modal interactions between visual and textual modalities~\cite{xu2024mh,moon2023query,moon2023correlation}, and leveraging prior knowledge from the large language models (LLMs) and event boundary cues~\cite{jiang2024prior,jang2023knowing,boris2024surprising}. Among existing architectures, the DETR-style structure~\cite{carion2020end}, which encodes visual and textual information into a set of learnable queries, has gained popularity~\cite{lei2021detecting}. In parallel, other methods adopt FPN-style models~\cite{lin2017feature} to produce multi-scale temporal feature sequences~\cite{yan2023unloc,mu2024snag,liu2024r}. Despite architectural differences, most prior works rely on frame- or clip-level (hereafter referred to as frame-level) features extracted by pre-trained visual encoders such as CLIP~\cite{radford2021learning} or I3D~\cite{carreira2017quo}, which primarily capture global semantics and overlook fine-grained object-level dynamics.

\begin{figure}[t]
    \centering
    \includegraphics[width=0.47\textwidth]{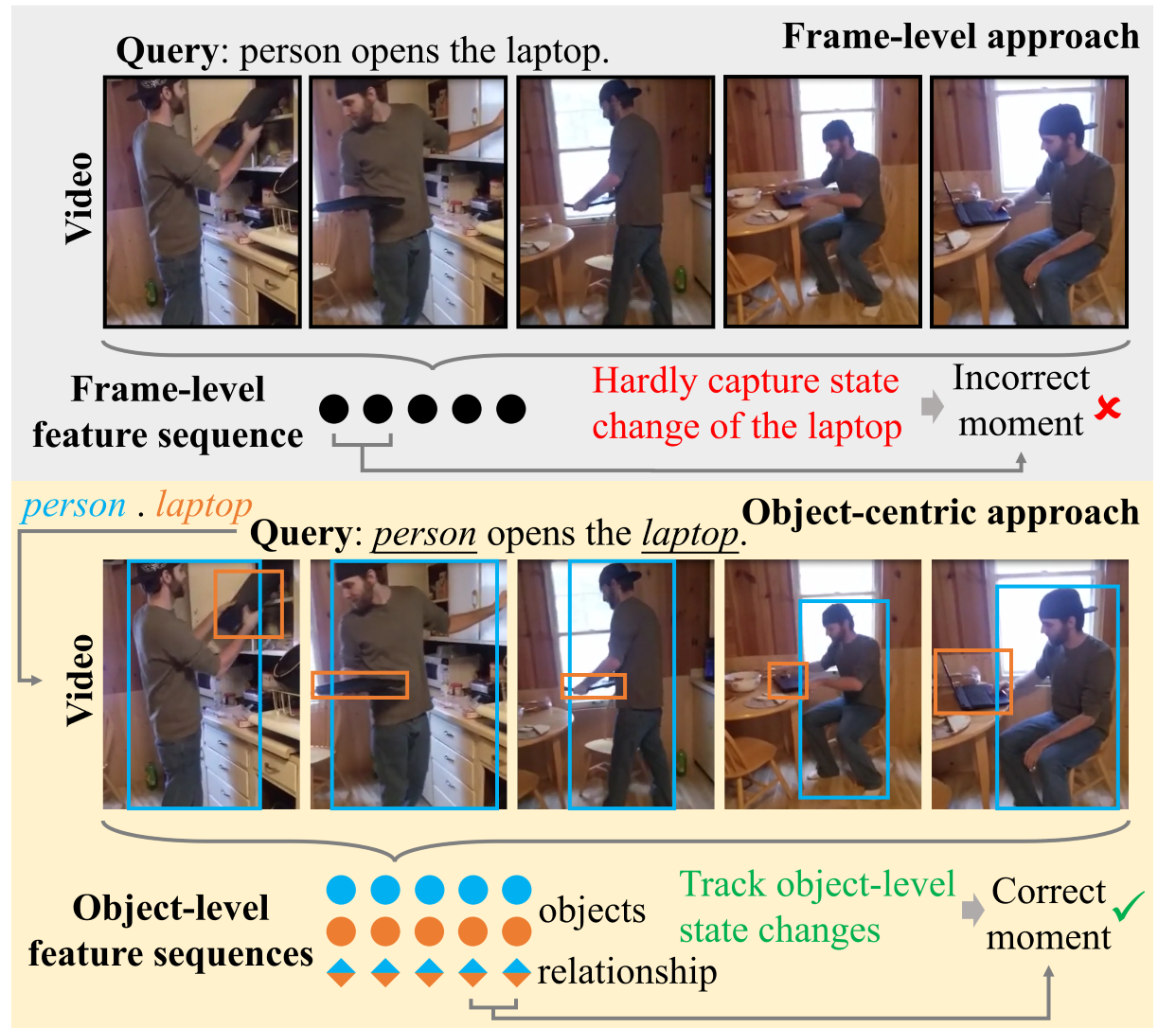}
    \caption{Comparison between the previous frame-level approach and the proposed object-centric approach. The frame-level approach fails to capture subtle yet critical state changes—such as the opening or closing of a laptop—resulting in inaccurate temporal localization. In contrast, the proposed object-centric approach explicitly tracks the states and interactions of relevant objects over time, enabling more precise moment localization.}
    \label{fig1}
\end{figure}

Despite these advancements, a fundamental question remains underexplored: \textit{Are frame-level features truly sufficient for video moment retrieval?} In many practical scenarios, queries are inherently object-oriented, often describing actions or events involving state changes at the object level. Nevertheless, such fine-grained queries have largely been addressed using object-agnostic approaches based on coarse frame-level representations. While powerful visual encoders may implicitly capture some object-level temporal cues, frame-level features typically lack explicit local information—such as object semantics, appearance, and inter-object relationships—that is critical for accurately localizing moments described by object-oriented queries. This limitation poses a significant barrier to further performance improvement, particularly in cases requiring precise modeling of object-level dynamics.

To ameliorate the aforementioned limitations, we argue that object-oriented queries should be handled in an object-centric manner, with temporal modeling performed explicitly at the object level. \figref{fig1} illustrates the contrast between our object-centric approach and the conventional frame-level approach. The example video depicts a person taking out a laptop, placing it on a table, and subsequently opening it. Given the query \textit{“person opens the laptop”}, the frame-level approach struggles to capture the crucial state transition of the laptop—from closed to open—due to its reliance on coarse frame-level features. This leads to inaccurate moment localization, especially since the laptop occupies only a small portion of the visual scene. In contrast, the object-centric approach detects and tracks the relevant objects (i.e., “person” and “laptop”) and predicts their relationships throughout the video. Based on this, it constructs object-level feature sequences that encode semantic and appearance information for each object. Similarly, for each pair of related objects, a relational feature sequence is constructed to track the evolution of their interaction. Temporal modeling over these object and relationship sequences enables the model to explicitly capture state transitions aligned with the query, offering a finer and more accurate understanding than modeling over frame-level features alone.

As previously stated, this paper proposes a novel object-centric moment retrieval framework. Specifically, the framework first extracts query-relevant objects by parsing the input query into a scene graph using a scene graph parser~\cite{li2023factual}. An open-vocabulary scene graph generation (OVSGG) model~\cite{chen2023expanding} is then employed to produce scene graphs for each video frame, along with associated object and relationship features. These features are organized into temporally ordered sequences using a tracking algorithm~\cite{zhang2022bytetrack}. To capture object-level state changes aligned with the query, a relational tracklet transformer is applied to the constructed object and relationship feature sequences. By modeling the spatio-temporal correlations among tracked objects and their interactions, the model enables more precise localization of moments corresponding to object-oriented queries.

The key contributions of this paper are as follows:
\begin{itemize}
    \item We propose a novel object-centric framework for video moment retrieval, explicitly designed to capture object-level state changes that are often overlooked by prior frame-level methods.
    \item Within this framework, we introduce a scene-graph-based object-level feature sequence representation that encodes rich semantic and visual information. In conjunction, we design a relational tracklet transformer to model spatio-temporal correlations among objects and their interactions.
    \item Extensive experiments on three benchmark datasets—Charades-STA, QVHighlights, and TACoS—demonstrate that our method consistently outperforms state-of-the-art methods, highlighting the importance of explicit object-level temporal modeling.
\end{itemize}


\section{Related Work}
\label{sec:2}

\begin{figure*}[t]
    \centering
    \includegraphics[width=\textwidth]{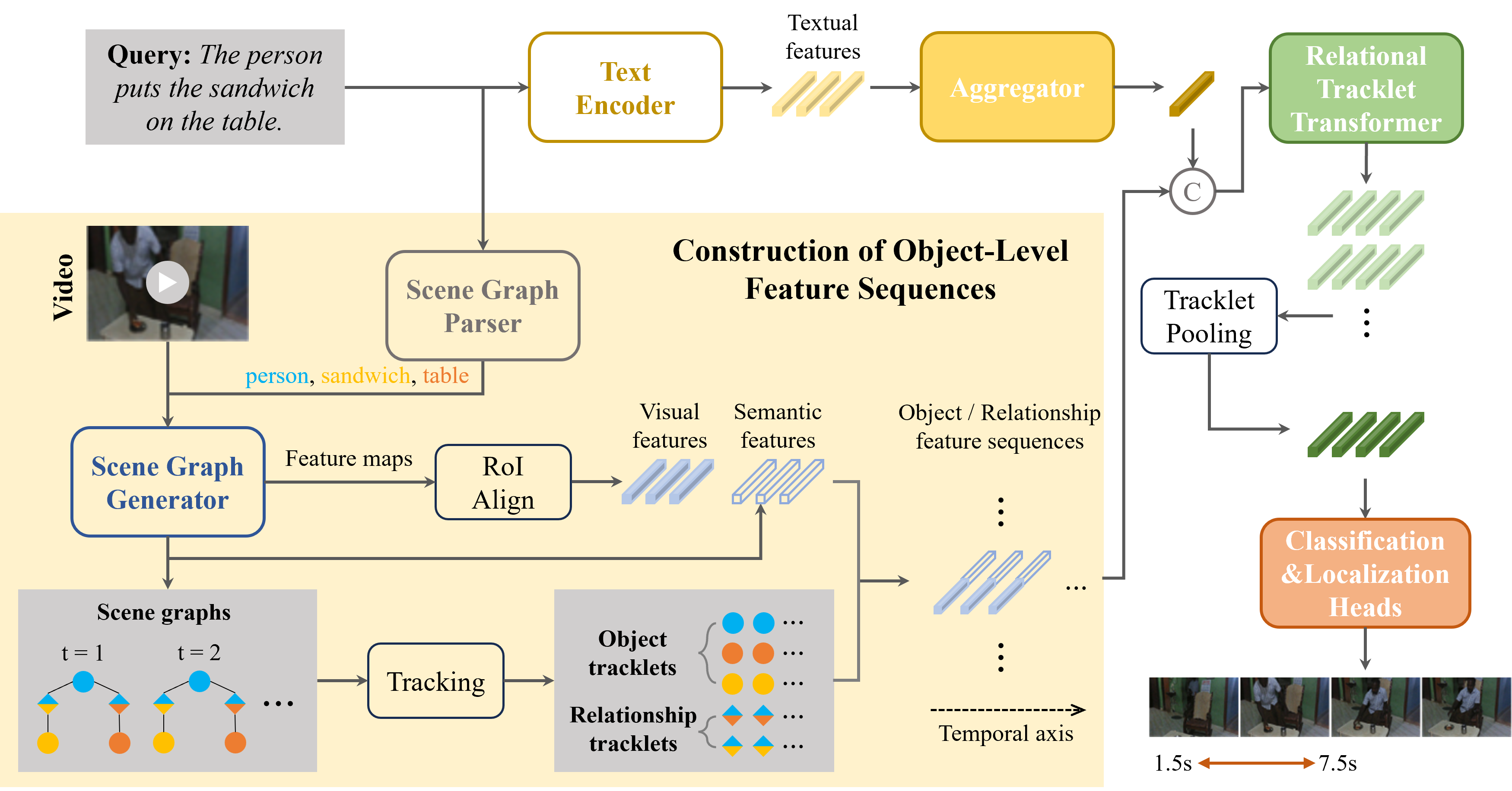}
    \caption{Overview of our object-centric moment retrieval framework. Given a video and a query, the framework first extracts query-relevant objects by parsing the query using a scene graph parser. It then constructs object-level feature sequences by embedding both visual and semantic information of these objects and their relationships into tracklets obtained from scene graph generation and object tracking. These feature sequences are concatenated with the query's textual feature and passed to a relational tracklet transformer, which models the spatio-temporal correlations among objects and relationships. The resulting representation enables accurate moment classification and localization by capturing query-relevant object state changes.}
    \label{fig2}
\end{figure*}

Existing video moment retrieval methods differ in candidate generation strategies and can be broadly categorized into proposal-based two-stage methods~\cite{gao2017tall,gao2021fast,soldan2021vlg,xiao2021boundary} and proposal-free one-stage methods~\cite{lu2019debug,mun2020local,zeng2020dense,zhang2020span}. Two-stage methods first generate temporal proposals and then evaluate their relevance and boundaries, whereas one-stage methods directly regress the relevant temporal segment. Due to their superior efficiency and the strong temporal modeling capabilities of transformers, recent efforts have primarily focused on one-stage paradigms.

Most recent methods adopt either a point-based or query-based framework. The point-based framework~\cite{yan2023unloc,mu2024snag,liu2024r} classifies temporal anchor points and regresses moment boundaries around them, drawing inspiration from point-based object detectors such as FCOS~\cite{tian2019fcos}. In contrast, the query-based framework~\cite{lei2021detecting,moon2023query,xu2024mh,lee2025bam}, inspired by DETR~\cite{carion2020end}, introduces learnable queries into a transformer to jointly encode visual and textual information for moment prediction. These DETR-style models are commonly used for both moment retrieval and highlight detection tasks, especially under the unified setting of the QVHighlights dataset~\cite{lei2021detecting}.

Several recent works extend moment retrieval beyond its traditional formulation, unifying it with other video understanding tasks such as temporal action localization~\cite{zeng2025unimd} and video summarization~\cite{lin2023univtg}, aiming for improved generalization. Some works focus on enhancing cross-modal interaction: QD-DETR~\cite{moon2023query} and MH-DETR~\cite{xu2024mh} investigate advanced cross-attention mechanisms, while EaTR~\cite{jang2023knowing} and LLMEPET~\cite{jiang2024prior} incorporate event boundary information to reduce distraction from irrelevant segments. Additionally, LLMEPET integrates an LLM encoder to refine cross-modal features. Unlike these methods, which typically adopt frozen backbones, R$^2$-Tuning~\cite{liu2024r} enables efficient transfer to video-language tasks by finetuning CLIP with a lightweight module.

MMRG~\cite{zeng2021multi} is most relevant to our work. It captures object-level state changes by constructing a multi-modal relational graph based on interactions among visual and textual objects. However, MMRG employs a simplified graph structure that focuses on a single entity for each object category extracted from the query. In contrast, our object-centric framework enables fine-grained modeling of spatio-temporal correlations across multiple objects, allowing better generalization to complex scenarios. Graph-based representations have also been explored in highlight detection and video QA. VH-GNN~\cite{zhang2020find} constructs spatial graphs over object nodes within frames and temporal graphs over frame nodes. However, as spatial graphs are aggregated independently into frames, it only models temporal dynamics at the frame level. Our method, by contrast, captures spatio-temporal correlations directly at the object level. VGT~\cite{xiao2022video} introduces a video graph for question answering, but is constrained to short clips and lacks the long-range modeling capacity required for moment retrieval. Moreover, unlike these prior graph-based methods, our framework leverages explicit relationship information from scene graphs, enabling a more precise understanding of object states and their temporal evolution.


\section{Object-Centric Framework for Moment Retrieval}
\label{sec:3}
Given a video $\mathbf{V}$ represented as a sequence of $L$ frames $[\mathbf{v}_1, \mathbf{v}_2, ..., \mathbf{v}_L]$, the goal of video moment retrieval is to identify the temporal segment that corresponds to a natural language query. \figref{fig2} presents an overview of the proposed object-centric framework. Unlike conventional approaches that rely solely on frame-level features, our method constructs feature sequences for both objects relevant to the query and their relationships. To model spatial and temporal correlations among these entities, we design a relational tracklet transformer that operates over the object-level feature sequences, enabling the precise localization of moments involving object-level state changes.

In Section~\ref{sec:3.1}, we introduce a frame-level baseline that serves as a comparative foundation. Section~\ref{sec:3.2} highlights the limitations of frame-level representations and motivates our object-centric approach. We then describe the construction of object-level inputs in Section~\ref{sec:3.3}, followed by the architecture of our object-centric model in Section~\ref{sec:3.4}.

\subsection{Baseline with Frame-Level Features}
\label{sec:3.1}
Following previous works~\cite{yan2023unloc,mu2024snag}, we implement a baseline that utilizes frame-level feature sequences processed by a transformer model. Given a video $\mathbf{V} = [\mathbf{v}_1, \mathbf{v}_2, ..., \mathbf{v}_L]$, a pre-trained visual encoder (e.g., CLIP~\cite{radford2021learning}, I3D~\cite{carreira2017quo}) is employed to extract a sequence of frame-level features $\mathbf{F} \in \mathbb{R}^{L \times d_{\mathbf{f}}}$, denoted as $[\mathbf{f}_1, \mathbf{f}_2, ..., \mathbf{f}_L]$. When using a clip-level encoder such as I3D, each $\mathbf{v}_i$ corresponds to a short video segment rather than a single frame.

In parallel, a pre-trained text encoder (e.g., CLIP~\cite{radford2021learning}, BERT~\cite{kenton2019bert}) encodes the input query into a sequence of token-level features $\mathbf{T} = [\mathbf{t}_1, \mathbf{t}_2, ..., \mathbf{t}_K] \in \mathbb{R}^{K \times d_{\mathbf{t}}}$. To enable multimodal fusion, we aggregate the token embeddings into a single textual feature vector $\bar{\mathbf{t}} \in \mathbb{R}^{d_{\mathbf{t}}}$ using a learnable linear attention mechanism. Specifically, a weight vector $\mathbi{w} \in \mathbb{R}^{d_{\mathbf{t}} \times 1}$ computes attention scores $\mathbi{a} = \mathbf{T} \cdot \mathbi{w}$, which are then used to derive the aggregated representation $\bar{\mathbf{t}} = \mathbi{a}^\top \cdot \mathbf{T}$ via weighted averaging. This aggregated feature is concatenated with each frame feature, yielding the fused visual-textual sequence $\tilbf{F}=[\tilbf{f}_1, \tilbf{f}_2, ..., \tilbf{f}_L]$

The sequence $\tilbf{F}$, augmented with temporal positional embeddings, is fed into a transformer consisting of stacked self-attention blocks. To handle moments of varying temporal durations, we generate multi-scale temporal features with resolutions $L$, $L/2$, $L/4$, etc., from intermediate transformer layers. Each element in the multi-scale sequences is treated as a moment candidate. A classification head predicts a relevance score $p$ for each candidate, while a localization head regresses the corresponding start and end times $(s, e)$. The top-$N$ scoring candidates $\{(p_i, s_i, e_i)\}_{i=1}^N$ are selected and post-processed using Soft-NMS~\cite{bodla2017soft} to suppress redundancy.

During training, the transformer and prediction heads are jointly optimized using Focal Loss~\cite{lin2017focal} for classification and Distance-IoU Loss~\cite{zheng2020distance} for localization. We adopt the Center Sampling strategy~\cite{zhang2022actionformer}, in which only candidates near ground-truth moment centers are considered positive. The localization loss is computed solely for these positive samples.

\subsection{Defects of Frame-Level Approaches}
\label{sec:3.2}
Most existing moment retrieval methods rely on frame-level features, assuming global representations suffice for aligning visual content with textual queries. However, this assumption often breaks down for object-oriented queries. We identify three core limitations:

\textbf{Lack of Fine-Grained Visual Detail.}
Frame-level features lack localized visual information. Pre-trained visual encoders are optimized for image- or video-level objectives, which prioritize global semantics over local object appearance. As a result, the final-layer features predominantly encode high-level concepts while overlooking subtle visual cues such as an object’s state (e.g., whether a laptop is open or closed). Since state changes may not be accompanied by semantic shifts, capturing such transitions requires fine-level visual detail, which frame-level features often miss.

\textbf{Insufficient Modeling of Object Relationships.}
Frame-level representations also fail to capture inter-object relationships, which are essential for reasoning about object states in context (e.g., “person opens laptop” implies physical interaction). This shortcoming arises because visual encoders are rarely trained to model structured relations among objects. Consequently, relational information critical for understanding events and actions is underrepresented.

\textbf{Inability to Track Multiple Objects.}
Finally, frame-level approaches struggle to maintain consistent representations of multiple objects over time. Even if the dominant object in a scene is adequately captured, other relevant entities may be spatially small or neglected due to attention dilution. As a result, modeling the dynamics of multiple objects becomes infeasible with frame-level features alone.

To overcome these limitations, we propose an object-centric framework that explicitly models object-level appearance and relational dynamics. By leveraging scene graph generation (SGG) to detect objects and relationships at both the visual and semantic levels, and organizing them into temporally coherent tracklets, our method provides fine-grained, structured inputs for downstream modeling.

\subsection{Object-Level Input Construction}
\label{sec:3.3}
As illustrated in \figref{fig2}, the construction of object-level inputs begins with parsing the query into a scene graph. We use a pre-trained scene graph parser~\cite{li2023factual}, implemented as a lightweight language model, to extract object nodes and their relationships from the natural language query. The resulting object class list is used to prompt an open-vocabulary scene graph generation (OVSGG) model~\cite{chen2023expanding}, which detects instances of the relevant objects and their pairwise relationships in each video frame. We adopt the OVSGG model pre-trained on the Visual Genome (VG) dataset~\cite{krishna2017visual}. In addition to the object prompt, the OVSGG model requires a relationship class list as input. Since state-of-the-art models are still unable to reliably detect open world relationships, we use the fixed set of relationship classes from the VG dataset rather than deriving them from the query.

The OVSGG model outputs detection results along with intermediate feature maps from its Swin Transformer~\cite{liu2021swin} backbone. We apply RoIAlign~\cite{he2017mask} to extract RoI features for each detected object or relationship. For relationships, RoI features are computed using the union of the subject and object bounding boxes. In parallel, we extract semantic features aligned with the text prompts from the final output of the OVSGG model. These visual and semantic features are complementary, together capturing both appearance-level and concept-level aspects of object states.

After detection, we apply a tracking algorithm~\cite{zhang2022bytetrack} to associate detected objects and relationships across time. Due to the low frame rate used in our setup (e.g., 1 or 0.5 fps), we perform feature-based matching using cosine similarity rather than IoU. The resulting object trajectories are organized into object tracklets, while relationships with confidence scores above a threshold are grouped into relationship tracklets, each corresponding to a specific object pair. Finally, we construct object-level feature sequences by embedding the concatenation of visual and semantic features into the corresponding tracklets. For time steps where a particular object or relationship is not present, we insert zero vectors to maintain temporal alignment.

\begin{figure}[t]
\centering
\includegraphics[width=0.47\textwidth]{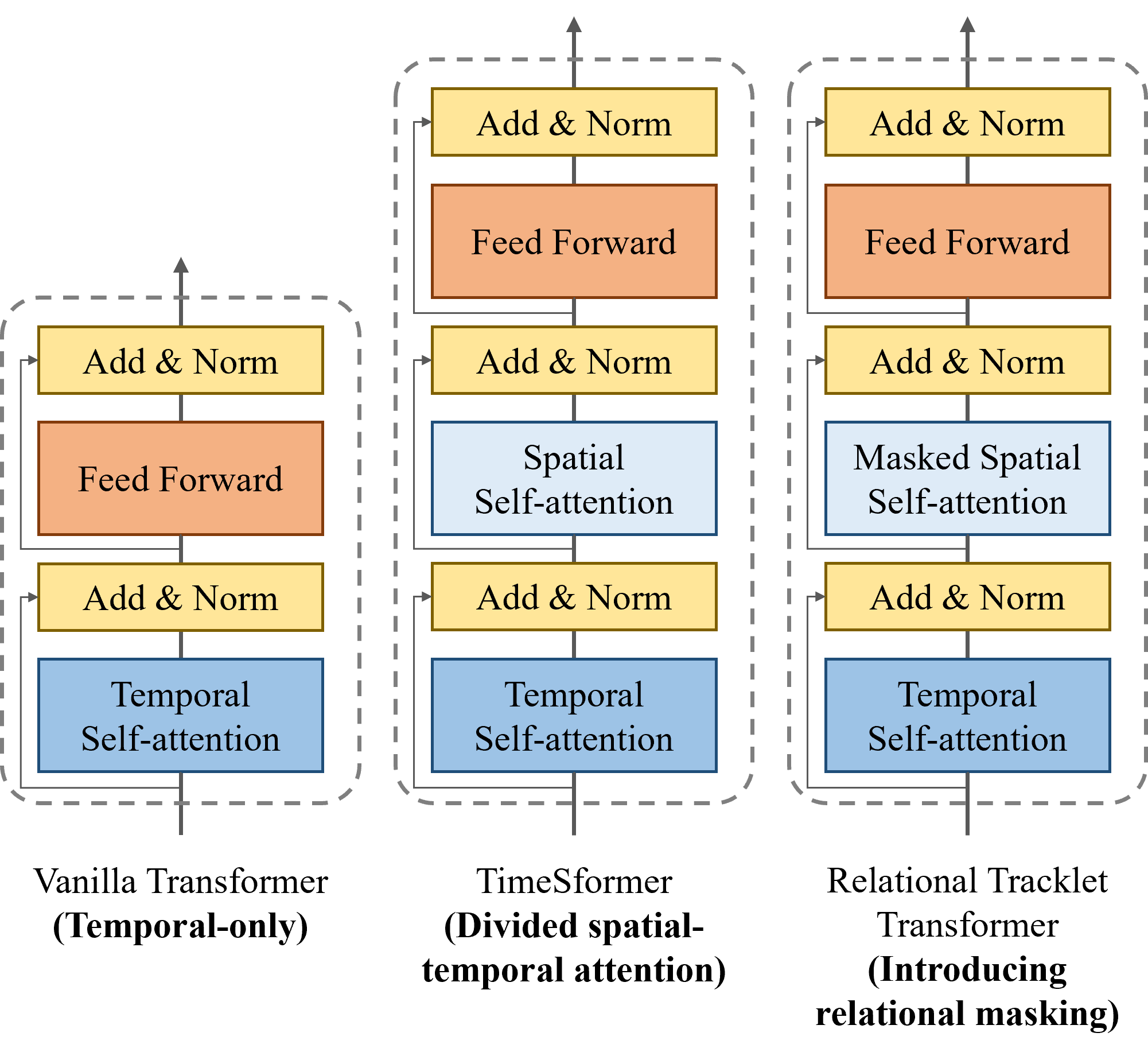}
\caption{Comparison between three transformer block variants. The proposed variant (right) integrates a relational mask, derived from scene graph information, into the spatial self-attention mechanism, enabling the model to attend more effectively to semantically relevant object pairs.}
\label{fig3}
\end{figure}

\subsection{Object-Centric Model Architecture}
\label{sec:3.4}
The object-level feature sequences introduced in Section~\ref{sec:3.3} require a model architecture capable of effectively capturing spatio-temporal correlations among objects and their interactions. To this end, we propose a relational tracklet transformer specifically designed to process temporally ordered object and relationship tracklets.

Our model builds upon the TimeSformer architecture~\cite{bertasius2021space}, which factorizes joint spatio-temporal self-attention into separate temporal and spatial attention modules, thereby improving efficiency. To better understand the differences, \figref{fig3} compares our relational tracklet transformer with a vanilla transformer and TimeSformer. A temporal-only vanilla transformer can be directly applied to object-level feature sequences but lacks the capacity to model spatial correlations. While TimeSformer addresses this by adding spatial attention, it treats all tracklets equally during spatial modeling, ignoring their semantic relationships.

To overcome this limitation, we introduce a \textit{graph-aware spatial attention mechanism}. Specifically, we construct a binary attention mask that reflects the underlying scene graph structure and restricts spatial attention to semantically related entities. The input to the transformer is a concatenated sequence of object and relationship tracklets: 
\[
[\mathcal{O}; \mathcal{R}] = [\mathbf{O}_1, \mathbf{O}_2, ..., \mathbf{O}_{N_{\mathcal{O}}}, \mathbf{R}_1, \mathbf{R}_2, ..., \mathbf{R}_{N_{\mathcal{R}}}],
\]
where $\mathcal{O}$ and $\mathcal{R}$ denote the sets of object and relationship feature sequences, respectively. A binary mask $\mathbi{m} \in \mathbb{R}^{(N_\mathcal{O}+N_\mathcal{R}) \times (N_\mathcal{O}+N_\mathcal{R})}$ is constructed such that $\mathbi{m}_{i,N_{\mathcal{O}}+j} = \mathbi{m}_{N_{\mathcal{O}}+j,i} = 1$ if object tracklet $\mathbf{O}_i$ is the subject or object of relationship tracklet $\mathbf{R}_j$, and 0 otherwise. Diagonal entries are set to 1 to preserve self-attention. This mask encodes the scene graph's relational structure and is applied during spatial self-attention.

More concretely, for each temporal position $j$, the spatial self-attention is computed over the set of feature vectors
\[
[\mathbf{o}_{1,j}, \mathbf{o}_{2,j}, ..., \mathbf{o}_{N_{\mathcal{O}},j}, \mathbf{r}_{1,j}, \mathbf{r}_{2,j}, ..., \mathbf{r}_{N_{\mathcal{R}},j}],
\]
where $\mathbf{o}_{i,j}$ and $\mathbf{r}_{i,j}$ are the object and relationship features at time $j$ from $\mathbf{O}_i$ and $\mathbf{R}_i$, respectively. In contrast, temporal self-attention is applied independently within each tracklet across time.

The textual feature aggregator and prediction heads in the object-centric model adopt the same design as those in the frame-level baseline (Section~\ref{sec:3.1}). Since the relational tracklet transformer produces one feature sequence per input tracklet, we apply max pooling along the tracklet dimension to obtain a unified temporal feature sequence. This sequence is then forwarded to the classification and localization heads. To mitigate potential detection failures and to leverage the complementary global context provided by frame-level features, we incorporate an optional parallel branch based on frame-level representations. The outputs from the object-centric and frame-level branches are fused via element-wise addition before being passed to the prediction heads. The training objectives remain consistent with those of the frame-level baseline.


\section{Experiments}
\label{sec:4}

\begin{table*}[t]
    \setlength{\tabcolsep}{1mm}
    \centering
    \begin{tabular}{lccccccccccccc}
        \toprule
        \multirow{3}{*}{Method} & \multicolumn{4}{c}{\textbf{Charades-STA}} & \multicolumn{4}{c}{\textbf{TACoS}} & \multicolumn{5}{c}{\textbf{QVHighlights}} \\
        \cmidrule(lr){2-5}
        \cmidrule(lr){6-9}
        \cmidrule(lr){10-14}
        & \multicolumn{3}{c}{R1} & \multirow{2}{*}{mIoU} & \multicolumn{3}{c}{R1} & \multirow{2}{*}{mIoU} & \multicolumn{2}{c}{R1} & \multicolumn{3}{c}{mAP}\\
        \cmidrule(lr){2-4}
        \cmidrule(lr){6-8}
        \cmidrule(lr){10-11}
        \cmidrule(lr){12-14}
        & @0.3 & @0.5 & @0.7 & & @0.3 & @0.5 & @0.7 & & @0.5 & @0.7 & @0.5 & @0.75 & Avg. \\
        \midrule
        MMRG \cite{zeng2021multi} & 71.6 & 44.3 & - & - & \underline{57.8} & 39.3 & - & - & - & - & - & - & - \\
        UnLoc \cite{yan2023unloc} & - & 60.8 & 38.4 & - & - & - & - & - & 66.1 & 46.7 & - & - & - \\
        UniVTG \cite{lin2023univtg} & 72.6 & 60.2 & 38.6 & 52.2 & 56.1 & 43.4 & 24.3 & 38.6 & 65.4 & 50.1 & 64.1 & 45.0 & 43.6 \\
        MH-DETR \cite{xu2024mh} & - & 56.4 & 36.0 & - & - & - & - & - & 60.1 & 42.5 & 60.8 & 38.1 & 38.4 \\
        QD-DETR \cite{moon2023query} & - & 57.3 & 32.6 & - & - & - & - & - & 62.4 & 45.0 & 62.5 & 39.9 & 39.9 \\
        CG-DETR \cite{moon2023correlation} & 70.4 & 58.4 & 36.3 & 50.1 & 54.4 & 39.5 & 23.4 & 37.4 & 65.4 & 48.4 & 64.5 & 42.8 & 42.9 \\
        UVCOM \cite{xiao2024bridging} & - & 59.3 & 36.6 & - & - & 36.4 & 23.3 & - & 63.6 & 47.5 & 63.4 & 42.7 & 43.2 \\
        LLMEPET \cite{jiang2024prior} & 70.9 & - & 36.5 & 50.3 & 52.7 & - & 22.8 & 36.6 & \underline{66.7} & 49.9 & 65.8 & 43.9 & 44.1 \\
        SnAG$^\ast$ \cite{mu2024snag} & 73.8 & 62.9 & 41.9 & 54.1 & 50.9 & 41.2 & 30.9 & 38.6 & - & - & - & - & - \\
        BAM-DETR \cite{lee2025bam} & 72.9 & 60.0 & 39.4 & 52.3 & 56.7 & 41.5 & 26.8 & 39.3 & 62.7 & 48.6 & 64.4 & 46.3 & 45.4 \\
        R$^2$-Tuning \cite{liu2024r} & 70.9 & 59.8 & 37.0 & 50.9 & 49.7 & 38.7 & 25.1 & 35.9 & \textbf{68.0} & 49.4 & \underline{69.0} & 47.6 & 46.2 \\
        \midrule
        Ours (w/o frame-level) & \underline{75.1} & \underline{64.6} & \underline{44.4} & \underline{55.7} & \textbf{58.5} & \textbf{47.7} & \textbf{34.1} & \textbf{43.1} & 65.1 & \underline{50.6} & 68.4 & \underline{47.8} & \underline{46.4} \\
        Ours (w/ frame-level) & \textbf{75.5} & \textbf{65.1} & \textbf{46.1} & \textbf{56.2} & 56.3 & \underline{45.2} & \underline{33.6} & \underline{41.9} & 66.2 & \textbf{52.3} & \textbf{70.1} & \textbf{50.8} & \textbf{49.2} \\
        \bottomrule
    \end{tabular}
    \caption{Experimental results on Charades-STA~\cite{gao2017tall}, TACoS~\cite{regneri2013grounding}, and QVHighlights (\textit{test} split)~\cite{lei2021detecting}. The best and second-best values are highlighted in bold and underlined, respectively. ``Ours (w/o frame-level)'' denotes the object-centric framework without the frame-level branch, while ``Ours (w/ frame-level)'' includes the integration of a frame-level branch. For the latter, frame-level features are extracted using a combination of CLIP-B and SlowFast, following the common practice adopted by most previous methods. $^\ast$For fair comparison, SnAG is re-evaluated on Charades-STA and TACoS using the same feature lengths as those used in our method and most of the previous methods.}
    \label{tab1}
\end{table*}

\subsection{Experiment Setup}
\label{sec:4.1}
We evaluate our method on three widely used benchmarks: Charades-STA~\cite{gao2017tall}, QVHighlights~\cite{lei2021detecting}, and TACoS~\cite{regneri2013grounding}. Charades-STA contains 6,672 videos ($\approx$30s each) and 16,128 text queries. QVHighlights comprises 10,148 videos and 10,310 queries, each possibly aligned with multiple moments. TACoS includes 127 long cooking videos (average length 4.8 minutes) and 18,818 query-moment pairs.

Video frames are extracted at 1 fps for Charades-STA and TACoS, and at 0.5 fps for QVHighlights. For Charades-STA and TACoS, we report Recall@1 (R1) at various temporal IoU (tIoU) thresholds and mean IoU (mIoU) between the top-1 prediction and ground truth. For QVHighlights, where queries can correspond to multiple segments, we follow prior works and use mean Average Precision (mAP) across tIoU thresholds from 0.5 to 0.95.

In all experiments, we use CLIP-B~\cite{radford2021learning} as the text encoder. For the frame-level visual encoder, we evaluate two settings: (1) a combination of SlowFast-R50~\cite{feichtenhofer2019slowfast} and CLIP-B’s visual branch, and (2) VideoMAEv2-b~\cite{wang2023videomae}. Additional implementation details are provided in Appendix~A.

\subsection{Comparison with State-of-the-Art Methods}
\label{sec:4.2}
\tabref{tab1} reports performance comparisons between our method and recent state-of-the-art methods across the three benchmarks. Most prior methods operate on frame-level features extracted using encoders such as CLIP~\cite{radford2021learning}, SlowFast~\cite{feichtenhofer2019slowfast}, I3D~\cite{carreira2017quo}, or C3D~\cite{tran2015learning}. Our object-centric framework alone outperforms existing methods on all datasets. When combined with the frame-level baseline (as described in Section~\ref{sec:3.4}), further improvements are observed on Charades-STA and QVHighlights, though a performance drop occurs on TACoS.

On Charades-STA, the combined model (“Ours (w/ frame-level)”) surpasses prior methods by at least 4.2\% in R1@0.7, 2.2\% in R1@0.5, and 1.7\% in R1@0.3. It also achieves a 2.1\% gain in mIoU, indicating better overall localization accuracy. On TACoS, our object-centric framework significantly outperforms prior methods, with margins of 3.2\% (R1@0.7), 4.3\% (R1@0.5), and 3.8\% (mIoU). However, integrating frame-level features results in degraded performance. We hypothesize that this is due to the limited visual variance in TACoS, which consists mostly of static-camera recordings. In such cases, frame-level features struggle to capture subtle changes in small or rare objects, failing to complement the object-centric representation. On QVHighlights, our method achieves the highest average mAP, leading the second-best method by 3.0\%. Although several prior methods outperform ours in R1@0.5, they lag behind in all other metrics.

\begin{table*}[t]
    \setlength{\tabcolsep}{1mm}
    \centering
    \begin{tabular}{C{8ex}cC{8ex}c C{6ex}C{6ex}C{6ex}C{6ex} C{6ex}C{6ex} C{6ex}C{6ex}C{6ex}}
        \toprule
        \multicolumn{2}{c}{Object-level features} & \multicolumn{2}{c}{Frame-level features} & \multicolumn{4}{c}{\textbf{Charades-STA}} & \multicolumn{5}{c}{\textbf{QVHighlights}} \\
        \cmidrule(lr){1-2}
        \cmidrule(lr){3-4}
        \cmidrule(lr){5-8}
        \cmidrule(lr){9-13}
        \multirow{2}{*}{Visual} & \multirow{2}{*}{Semantic} & \multirow{2}{*}{SF+C} & \multirow{2}{*}{VideoMAEv2} & \multicolumn{3}{c}{R1} & \multirow{2}{*}{mIoU} & \multicolumn{2}{c}{R1} & \multicolumn{3}{c}{mAP} \\
        \cmidrule(lr){5-7}
        \cmidrule(lr){9-10}
        \cmidrule(lr){11-13}
        & & & & @0.3 & @0.5 & @0.7 & & @0.5 & @0.7 & @0.5 & @0.75 & Avg. \\
        \midrule
        & & \checkmark & & 72.5 & 61.8 & 43.1 & 53.4 & 59.4 & 46.1 & 63.8 & 45.4 & 44.1 \\
        & & & \checkmark & 79.1 & 69.5 & 52.1 & 59.9 & 60.7 & 47.6 & 65.2 & 47.9 & 46.1 \\
        \midrule
        \checkmark & & & & 72.5 & 61.8 & 42.0 & 53.4 & 61.8 & 48.5 & 65.2 & 46.6 & 44.8 \\
        & \checkmark & & & 72.6 & 60.9 & 39.4 & 52.3 & 61.0 & 47.6 & 64.8 & 45.7 & 44.1 \\
        \checkmark & \checkmark & & & 75.1 & 64.6 & 44.4 & 55.7 & 66.1 & 51.4 & 68.5 & 49.1 & 47.1 \\
        \midrule
        \checkmark & \checkmark & \checkmark & & 75.5 & 65.1 & 46.1 & 56.2 & 67.4 & 54.9 & 69.7 & 52.1 & 49.9 \\
        \checkmark & \checkmark & & \checkmark & 80.9 & 70.3 & 52.9 & 61.3 & 67.7 & 55.7 & 69.7 & 52.4 & 50.8 \\
        \bottomrule
    \end{tabular}
    \caption{Ablation study on object-level and frame-level features on Charades-STA and QVHighlights (\textit{val} split). Frame-level features are extracted using two settings: (1) a combination of SlowFast and CLIP-B (SF+C), and (2) VideoMAEv2-b.}
    \label{tab2}
\end{table*}

\subsection{Analysis of Feature Efficacy}
\label{sec:4.3}
To assess the impact of object-level and frame-level features, we conduct ablation studies summarized in \tabref{tab2}. In addition to SlowFast+CLIP-B (SF+C), we also evaluate VideoMAEv2-b, a powerful video encoder capable of modeling motion through multi-frame inputs.

The first two rows in \tabref{tab2} show results for the frame-level baseline. VideoMAEv2 consistently outperforms SF+C, especially on Charades-STA, where motion understanding is critical. The smaller gap on QVHighlights reflects its greater emphasis on object semantics rather than motion. Results in the next three rows demonstrate the importance of both visual and semantic features in the object-centric framework: removing either component leads to noticeable drops in performance, confirming their complementary roles in modeling object state changes.

The final two rows evaluate the combined model (object-centric + frame-level), which achieves consistent improvements over the object-centric model alone on both datasets. Notably, on Charades-STA, adding VideoMAEv2 features boosts mIoU by 5.6\%, compared to only 0.5\% with SF+C. In contrast, QVHighlights shows comparable mAP gains (2.8\% vs. 3.7\%).

Comparing the last two rows to the first two highlights the overall value of the object-centric framework. On Charades-STA, incorporating it yields mIoU improvements of 2.8\% (SF+C) and 1.4\% (VideoMAEv2). The smaller gain for VideoMAEv2 reflects its strong performance in person-centric videos, where motion modeling alone suffices. On QVHighlights, however, the object-centric framework consistently enhances both baselines, with mAP (Avg.) increases of 5.8\% and 4.7\%, further validating its contribution.

\begin{table}[t]
    \setlength{\tabcolsep}{1mm}
    \centering
    \begin{tabular}{ccccccccc}
        \toprule
        \multicolumn{2}{c}{Visual encoder} & \multicolumn{3}{c}{Scene graph} & \multicolumn{3}{c}{R1} & \multirow{2}{*}{mIoU} \\
        \cmidrule(lr){1-2}
        \cmidrule(lr){3-5}
        \cmidrule(lr){6-8}
        SF+C & VMAE & VG & AG & GT & @0.3 & @0.5 & @0.7 & \\
        \midrule
        & & \checkmark & & & 75.1 & 64.6 & 44.4 & 55.7 \\
        & & & \checkmark & & 75.4 & 65.2 & 45.2 & 56.0 \\
        & & & & \checkmark & 86.6 & 78.4 & 59.9 & 67.4 \\
        \midrule
        \checkmark & & \checkmark & & & 75.5 & 65.1 & 46.1 & 56.2 \\
        \checkmark & & & \checkmark & & 76.4 & 66.4 & 47.4 & 57.1 \\
        \checkmark & & & & \checkmark & 87.6 & 80.3 & 63.6 & 69.1 \\
        \midrule
        & \checkmark & \checkmark & & & 80.9 & 70.3 & 52.9 & 61.3 \\
        & \checkmark & & \checkmark & & 80.8 & 72.0 & 54.4 & 61.8 \\
        & \checkmark & & & \checkmark & 88.0 & 81.1 & 64.9 & 69.8 \\
        \bottomrule
    \end{tabular}
    \caption{Impact of scene graph quality on performance on Charades-STA. We evaluate the effect of different scene graph sources: pre-trained models on Visual Genome (VG) and Action Genome (AG), as well as ground-truth (GT) scene graphs. Frame-level features are extracted using two settings: (1) a combination of SlowFast and CLIP-B (SF+C), and (2) VideoMAEv2-b (VMAE).}
    \label{tab3}
\end{table}

\subsection{Impact of Scene Graph Quality}
\label{sec:4.4}
Since the object-centric framework relies on scene graphs, we investigate how scene graph quality affects performance. We train an OVSGG model on the Action Genome (AG) dataset~\cite{ji2020action}, which provides scene graph annotations for keyframes in Charades-STA. This model achieves better accuracy on Charades-STA than the Visual Genome (VG)-trained version and produces more reliable scene graphs. We also experiment with ground-truth (GT) scene graphs from AG to approximate the upper bound, although they are temporally sparse and limited in vocabulary.

\tabref{tab3} compares results using scene graphs from three sources: VG, AG, and GT. Without frame-level features, using AG and GT graphs yields mIoU improvements of 0.3\% and 11.7\%, respectively, over VG. When frame-level features are included, consistent gains are observed: +0.9\% and +12.9\% for SF+C, and +0.5\% and +8.5\% for VideoMAEv2. These results highlight the strong potential of our framework to further improve as better scene graph generation models become available.

\begin{table}[t]
    \setlength{\tabcolsep}{1mm}
    \centering
    \begin{tabular}{lcc}
        \toprule
        Module & \makecell{Time cost \\ (seconds/video)} & \makecell{GPU memory \\ usage (Mb)} \\
        \midrule
        Scene graph parser              & 0.142 & 1549 \\
        Text encoder                    & 0.005 & 885 \\
        Visual encoder                  & 2.290 & 2886 \\
        Scene graph generator           & 6.321 & 3413 \\
        Tracklet transformer & 0.115 & 1932 \\
        \bottomrule
    \end{tabular}
    \caption{Computational cost of major modules during inference on Charades-STA (measured on a single NVIDIA L40S GPU). The scene graph parser is FACTUAL (Flan-T5-base); the visual encoder is a combination of CLIP-B and SlowFast-R50; the text encoder is CLIP-B; and the scene graph generator is OvSGTR (Swin-B). The cost of the tracking process and the RoIAlign operation is negligible and accounted for in the runtime of the relational tracklet transformer and scene graph generator, respectively.}
    \label{tab4}
\end{table}

\subsection{Computational Cost Analysis}
\label{sec:4.5}
The object-centric framework introduces additional components beyond frame-level methods, increasing computational cost. We analyze inference-time processing time and GPU memory usage of each major module on Charades-STA, as summarized in \tabref{tab4}. The main source of overhead lies in the scene graph generator, which consumes more GPU memory than the visual encoder but keeps total usage around 10.7 GB—acceptable for most practical settings. In terms of processing time, the scene graph generator is about 2.76$\times$ slower than the visual encoder, mainly due to its reliance on a heavy object detector. We expect this overhead can be mitigated by future OVSGG models based on real-time detectors such as YOLO-World~\cite{cheng2024yolo}, which are significantly faster than GroundingDINO~\cite{liu2024grounding}, used in our current setup. The costs of previous methods~\cite{lee2025bam,jiang2024prior} mainly come from the visual encoder which is the same as ours, while their transformer modules are generally less costly. Therefore, \tabref{tab4} allows a rough cost comparison between previous methods and our method.

\section{Conclusion}
\label{sec:5}

We proposed a novel object-centric framework for video moment retrieval that addresses the limitations of prior frame-level approaches in modeling object-level state changes. Our method constructs object-level feature sequences and captures their spatio-temporal dynamics via a relational tracklet transformer. Experiments on three benchmarks demonstrate that our method outperforms state-of-the-art methods. Additionally, we show that the framework's performance can be further improved as open-vocabulary scene graph generation models continue to advance.

\bibliography{aaai2026}

\end{document}


\maketitle
\appendix

\section{Implementation Details}
\subsection{Object-level Feature Sequence Construction}
We adopt a Flan-T5-base model trained on the FACTUAL dataset~\cite{li2023factual} as our scene graph parser. For scene graph generation, we use OvSGTR~\cite{chen2023expanding} with a Swin-B backbone pretrained on Visual Genome. The confidence thresholds for object and relationship detection are set to 0.2 and 0.3, respectively.

Visual features are extracted via RoIAlign (7$\times$7 bins) on the final feature map of the Swin-B backbone and aggregated through global average pooling. Semantic features aligned with the prompt embeddings are also extracted from OvSGTR. ByteTrack~\cite{zhang2022bytetrack} is used to track detected objects across frames. Given the low frame rate, we compute cosine similarity between both visual and semantic features for association, and average the two scores. Tracking hyperparameters are set as follows: \texttt{track\_thresh} = 0, \texttt{track\_buffer} = 30, and \texttt{match\_thresh} = 0.9.

\subsection{Model Details}
Our relational tracklet transformer architecture builds upon ActionFormer~\cite{zhang2022actionformer}, replacing its transformer blocks with our proposed design. The network consists of a feature embedding module and seven transformer blocks. Textual, object, and relationship features are first projected into a shared 256-dimensional space via separate linear layers. These are then concatenated and passed through two 1D convolutional layers (kernel size = 3). The resulting sequences are fed into the transformer stack, where temporal downsampling is achieved via strided attention (stride = 2) in several blocks—five for Charades-STA and QVHighlights, and six for TACoS.

Feature sequences at all temporal scales are forwarded to both classification and localization heads, each consisting of three 1D convolutional layers (kernel size = 3). Ground truth moments are assigned to temporal scales based on the moment duration, following ActionFormer.

\subsection{Training Details}
All models are trained on a single NVIDIA L40S GPU (48GB) using PyTorch. Results are reported from a single training run. We use SGD with a maximum learning rate of 1e-3 and a batch size of 16 for 10 epochs. The learning rate schedule includes a linear warm-up over the first 3 epochs, followed by cosine annealing.

Losses include the Focal loss and Distance-IoU loss, weighted equally. For QVHighlights, we additionally apply a saliency loss, as in~\cite{xu2024mh}, using the provided highlight annotations. The saliency loss is weighted equally with the others.

\begin{figure*}[t]
    \centering
    \includegraphics[width=\textwidth]{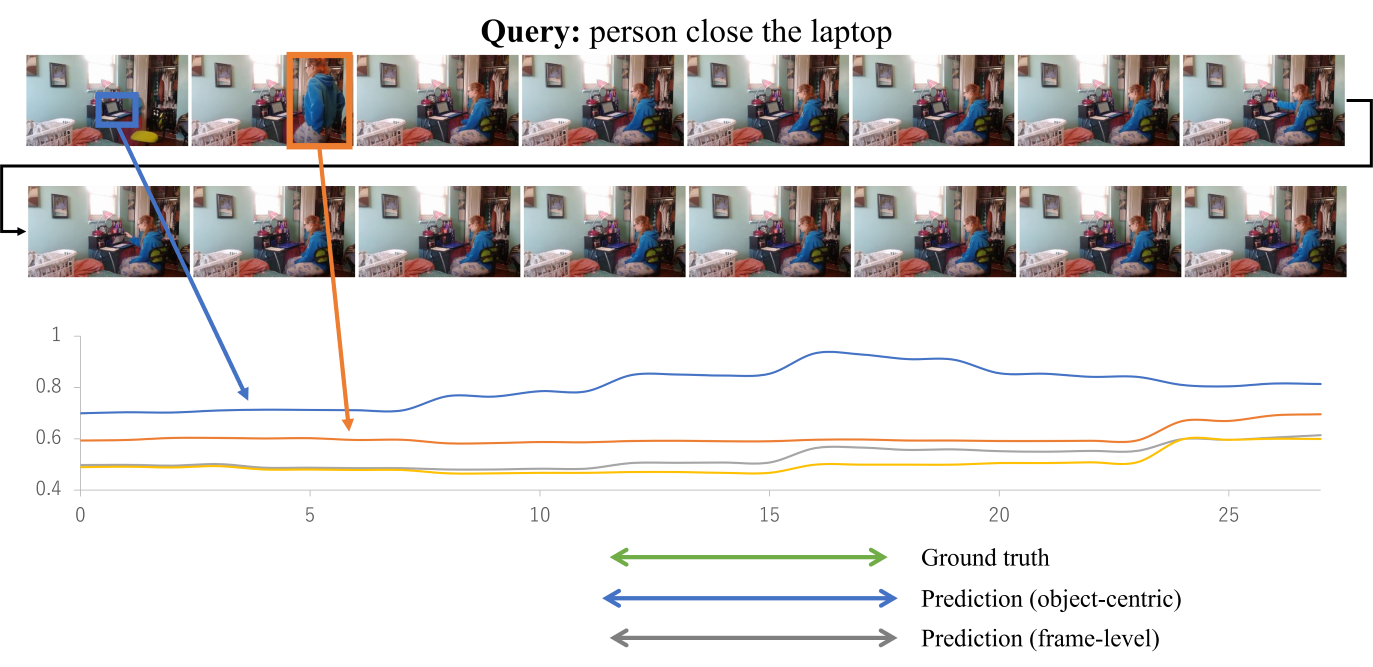}
    \caption{Illustration of retrieval results alongside object feature visualizations. The line graph shows the mean activation values of object features over time, where the x-axis denotes time (in seconds) and the y-axis indicates the feature magnitude. Each curve corresponds to a specific object, revealing its temporal relevance to the query.}
    \label{sup_fig1}
\end{figure*}

\begin{figure*}[t]
    \centering
    \includegraphics[width=\textwidth]{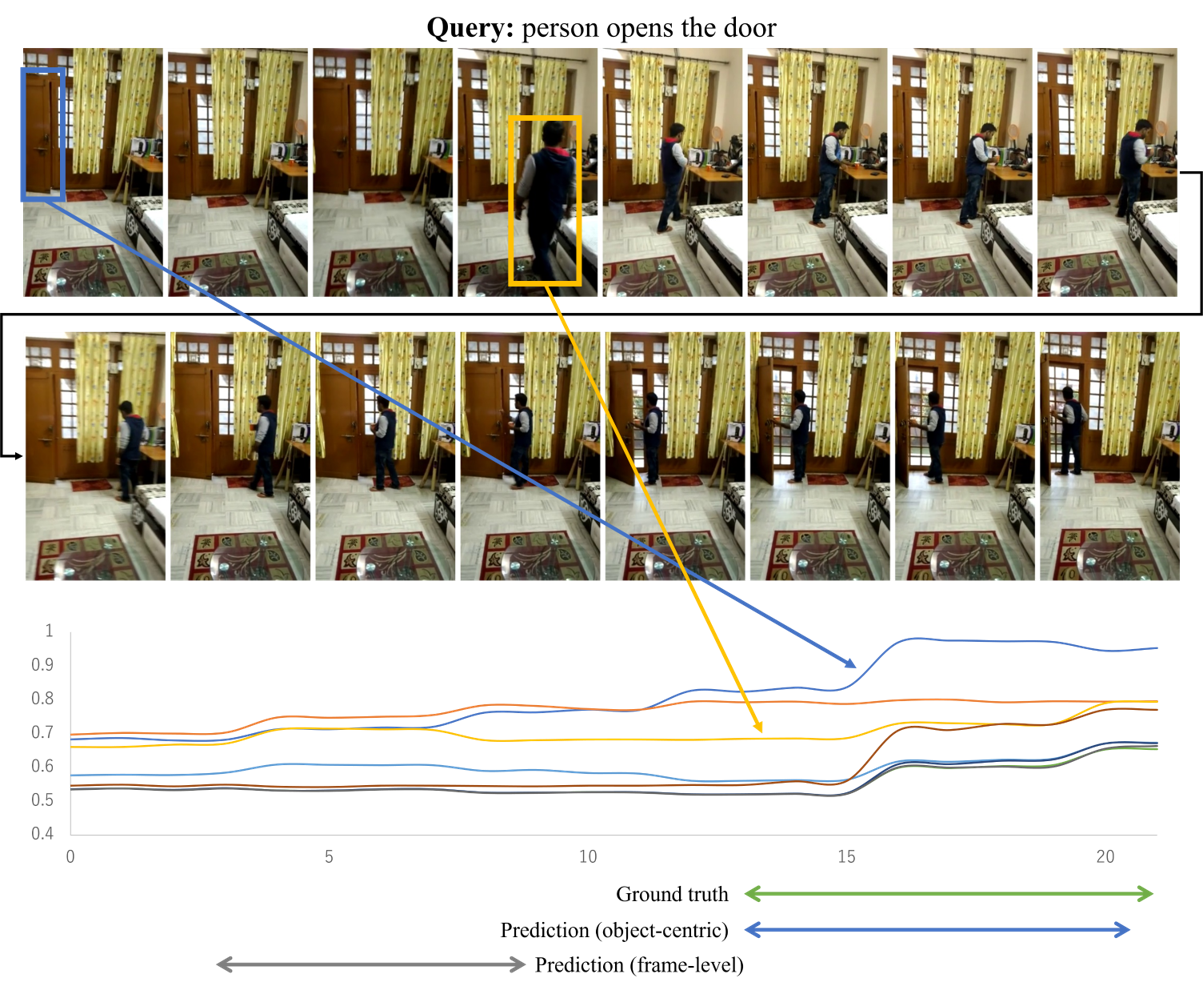}
    \caption{Illustration of retrieval results alongside object feature visualizations. The line graph shows the mean activation values of object features over time, where the x-axis denotes time (in seconds) and the y-axis indicates the feature magnitude. Each curve corresponds to a specific object, revealing its temporal relevance to the query.}
    \label{sup_fig2}
\end{figure*}

\begin{figure*}[t]
    \centering
    \includegraphics[width=\textwidth]{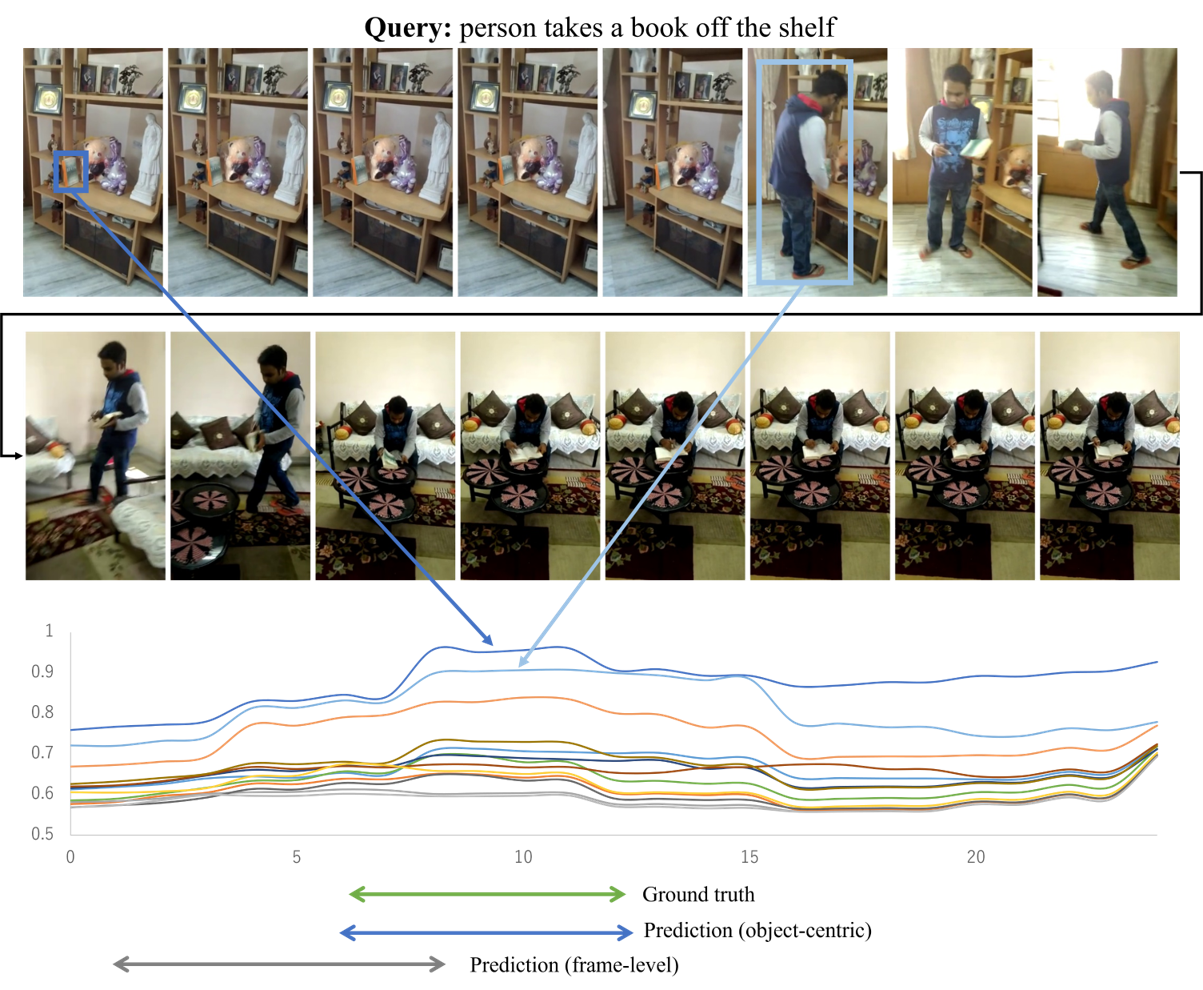}
    \caption{Illustration of retrieval results alongside object feature visualizations. The line graph shows the mean activation values of object features over time, where the x-axis denotes time (in seconds) and the y-axis indicates the feature magnitude. Each curve corresponds to a specific object, revealing its temporal relevance to the query.}
    \label{sup_fig3}
\end{figure*}

\section{Visualization}
To qualitatively assess the advantages of our object-centric framework over the frame-level baseline, we visualize three examples in \figref{sup_fig1}, \figref{sup_fig2}, and \figref{sup_fig3}. In each example, we also plot the temporal activation curves derived from the object features, which are output by the relational tracklet transformer. Specifically, we compute the mean activation of each object’s feature at every time step, yielding a 1D temporal curve per object. These curves reflect the relative relevance of each object to the query over time, offering insight into how the model attends to different entities.

In \figref{sup_fig1}, both methods successfully localize the moment corresponding to the query. While the frame-level baseline may lack the granularity to capture the subtle state transition of the laptop from open to closed, it appears to rely on detecting the person performing the action. In contrast, the object-centric model explicitly attends to the laptop and effectively associates its state change with the query, as evidenced by a spike in its activation within the relevant moment.

In \figref{sup_fig2}, only the object-centric framework correctly identifies the relevant moment. Here, the state change of the door is visually salient, yet the frame-level baseline fails possibly due to camera motion. In contrast, our method captures a distinct peak in the door activation curve during the opening action, leading to accurate localization.

In \figref{sup_fig3}, the baseline is distracted by the visual dominance of the shelf, which occupies much of the frame. Consequently, it predicts the moment incorrectly. On the other hand, the object-centric framework selectively focuses on the book and person, whose feature activations rise sharply at the correct time, resulting in a more precise prediction.

Across all three examples, we observe that the object-centric framework consistently highlights the most relevant objects by their elevated feature values around the ground truth moment. These objects also maintain higher activations than irrelevant ones. This supports our central claim: explicitly modeling object-level dynamics enables the model to focus on query-relevant entities and their state changes, resulting in more accurate temporal grounding. In contrast, the frame-level baseline often fails under background clutter or motion noise, due to its limited granularity and lack of structured object awareness.

\bibliography{aaai2026}